\documentclass{article}

 \usepackage[final]{nips_2016}

\usepackage[utf8]{inputenc} % allow utf-8 input
\usepackage[T1]{fontenc}    % use 8-bit T1 fonts
\usepackage{hyperref}       % hyperlinks
\usepackage{url}            % simple URL typesetting
\usepackage{booktabs}       % professional-quality tables
\usepackage{amsfonts}       % blackboard math symbols
\usepackage{nicefrac}       % compact symbols for 1/2, etc.
\usepackage{microtype}      % microtypography

\usepackage{amssymb}
\usepackage{algorithm}
\usepackage{algorithmic}
\usepackage{graphicx} % more modern
\usepackage{subfigure}
\usepackage{wrapfig}

\newtheorem{define}{Definition}

\newtheorem{theorem}{Theorem}

\newcommand{\Algorithm}{Iterative Hierarchical Optimization for Misspecified Problems}
\newcommand{\Alg}{IHOMP}

\newcommand{\MProb}{MP}

\title{Iterative Hierarchical Optimization for Misspecified Problems (IHOMP)}

\author{
  Daniel J.~Mankowitz \\
  Electrical Engineering Department,\\
   The Technion - Israel Institute of Technology,\\
    Haifa 32000, Israel\\
  \texttt{danielm@tx.technion.ac.il} \\
   \And
   Timothy A.~Mann \\
   Google Deepmind \\
   London, UK \\
   \texttt{timothymann@google.com} \\
   \AND
   Shie Mannor \\
   Electrical Engineering Department,\\
   The Technion - Israel Institute of Technology,\\
    Haifa 32000, Israel\\
   \texttt{shie@ee.technion.ac.il} \\
}

\begin{document}
% \nipsfinalcopy is no longer used

\maketitle

\begin{abstract}
  For complex, high-dimensional Markov Decision Processes (MDPs), it may be necessary to represent the policy with function approximation. A problem is misspecified whenever, the representation cannot express any policy with acceptable performance. We introduce \Alg\ : an approach for solving misspecified problems. \Alg\ iteratively learns a set of context specialized options and combines these options to solve an otherwise misspecified problem. Our main contribution is proving that \Alg\ enjoys theoretical convergence guarantees. In addition, we extend \Alg\ to exploit Option Interruption (OI) enabling it to decide where the learned options can be reused. Our experiments demonstrate that \Alg\ can find near-optimal solutions to otherwise misspecified problems and that OI can further improve the solutions.
\end{abstract}

%%%%%%%%%%%%%%%%%%%%%%%%%%%%%%%%%%%%%%%%%%%%%%%%%%%%%%%%%%%%%%%%%%%%%%%%%%%%%%%
%
% INTRO
%
%%%%%%%%%%%%%%%%%%%%%%%%%%%%%%%%%%%%%%%%%%%%%%%%%%%%%%%%%%%%%%%%%%%%%%%%%%%%%%%
\section{Introduction}
\label{sec:intro}

%What is the problem?
Reinforcement Learning (RL) algorithms can learn near-optimal solutions to well-defined problems. However, real-world problems rarely come in the form of a concrete problem description. A human has to translate the poorly defined target problem into a concrete problem description. A Misspecified Problem (\MProb) occurs when an optimal solution to the problem description is inadequate in the target problem. Unfortunately, creating a well-defined problem description is a challenging art. Furthermore, \MProb s can have serious consequences in many domains ranging from smart-grids \citep{Abiri2013,Wu2010} and robotics \citep{Smart2002} to inventory management systems \citep{Mann2014a}. In this paper, we introduce a hierarchical approach that mitigates the consequences of problem misspecification.

%What type of misspecified problems do we consider?
RL problems are often described as Markov Decision Processes \citep[MDPs]{Sutton1998}. A solution to a MDP is a function that generates an action when presented with the current state, called a \textit{policy}. The solution to a MDP is any policy that maximizes the long term sum of rewards. For problems with continuous, high dimensional state-spaces, explicitly representing the policy is infeasible, thus for the remainder of this paper we restrict our discussion to linearly parametrized policy representations \citep{Sutton1996,Roy2013}.\footnote{Our results are generalizable and complementary to non-linear parametric policy representations.}

{\bf Why are problems misspecified?} While a problem description can be misspecified for many reasons, one important case is due to the state representation. It is well established in the machine learning \citep{Levi2004,Zhou2009} and RL \citep{Konidaris2011} literature that ``good'' features can have a dramatic impact on performance. Finding ``good'' features to represent the state is a challenging domain specific problem that is generally considered outside of the scope of RL. Unfortunately, domain experts may not supply useful features either because they do not fully understand the target problem or the technicalities of reinforcement learning.

In addition, we may prefer a \MProb\ with a limited state representation for several reasons: (1) \textit{Regularization}: We wish to have a limited feature representation to improve the generalization and avoid overfitting \citep{Singh1995,Geramifard2012}. (2) \textit{Memory and System constraints}: Only a finite number of the features can be used due to computational constraints \citep{Roy2013,Singh1995}. In real-time systems, querying a feature may take too long. In physical systems, the sensor required to measure a desired feature may be prohibitively expensive. (3) \textit{Learning on Large Data}: After learning on large amounts of data, augmenting a feature set with new features to get improved performance is non-trivial and often inefficient \citep{Geramifard2012}.

%What has been done to deal with model misspecification
{\bf How can we mitigate misspecification?} Learning a {\it hierarchical policy} can mitigate the problems associated with a \MProb\ and contrast this against a {\it flat policy} approach where a single, parameterized policy is used to solve the entire MDP.

\begin{figure*}
\centering
\includegraphics[width=0.85\textwidth]{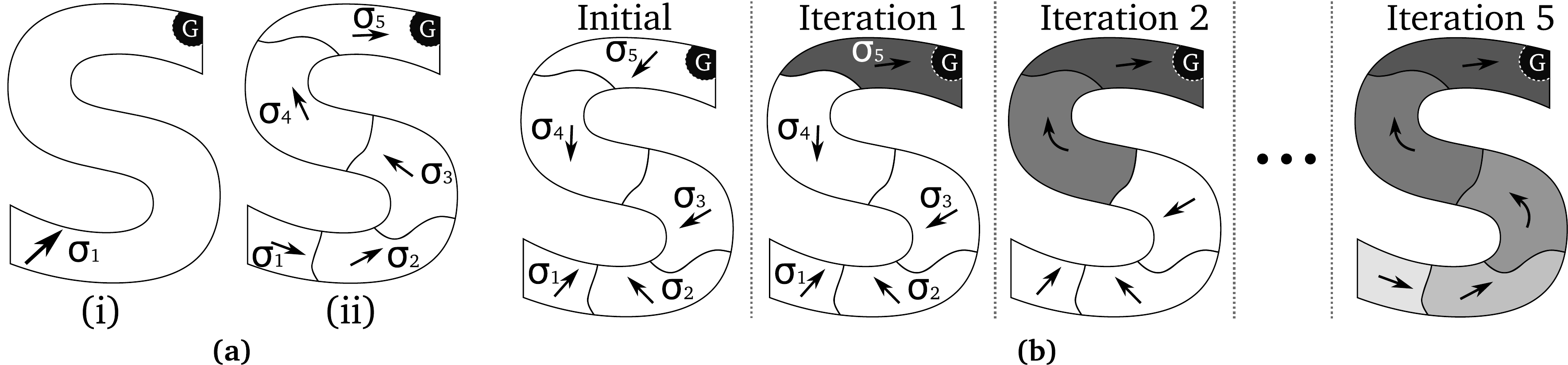}
\caption{($a$) An episodic MDP with S-shaped state-space and goal region $G$. ($i$) Flat approach: A single policy failing to solve the entire task resulting in a misspecified model ($ii$) Hierarchical approach: Using and combining simple policy representations to solve a task. ($b$) Learning options, denoted by black arrows. The domain is partitioned into five classes (sub-partitions) resulting in option-set $\Sigma = \{ \sigma_1, \sigma_2, \sigma_3, \sigma_4, \sigma_5 \}$. In iteration $1$, all options except for $\sigma_5$ (which has immediate access to the goal region) are arbitrary. In iteration $2$, $\sigma_5$ propagates reward back to $\sigma_4$. This process repeats until useful options are learned over the entire state-space.}
\label{fig:compact}
\end{figure*}

To illustrate how learning a hierarchical policy can repair MPs, consider the S-shaped domain shown in Figure \ref{fig:compact}$a$. To solve the task the agent must move from the bottom left corner to the goal region denoted by the letter `G' in the top right. The state representation only permits policies that move in a straight line. So the problem is misspecified, and it is not solvable with a flat policy approach (Figure \ref{fig:compact}$a.i$). However, if we break up the state-space, as shown in Figure \ref{fig:compact}$a.ii$, and learn one policy for each cell, the problem is solvable.

The partial policies shown in Figure \ref{fig:compact}$a.ii$ are an example of abstract actions, called options \citep{Sutton1999}, macro-actions \citep{Hauskrecht1998,He2011}, or skills \citep{Konidaris2009}. Learning useful options has been a topic of intense research \citep{McGovern2001,Moerman2009,Konidaris2009,Brunskill2014,Hauskrecht1998}. However, previous approaches have proposed algorithms for learning options to learn or plan faster. In contrast, our objective is to learn options to repair a MP.

{\bf Proposed Algorithm:} We introduce a meta-algorithm, \Algorithm\ (\Alg), that uses an RL algorithm as a ``black box'' to iteratively learn options that repair MPs. To force the options to specialize, \Alg\ uses a partition of the state-space and trains one option for each class in the partition (Figure \ref{fig:compact}$b$). Any arbitrary partitioning scheme can be used, however the partition impacts performance. During an iteration of \Alg, an RL algorithm updates each option. The options may be initialized arbitrarily, but after the first iteration options with access to a goal region or non-zero rewards will learn how to exploit those rewards (e.g., Figure \ref{fig:compact}$b$, Iteration 1). On further iterations, the newly acquired options propagate reward back to other regions of the state-space. Thus, options that previously had no reward signal exploit the rewards of other options that have received meaningful reward signals (e.g., Figure \ref{fig:compact}$b$, Iterations 2 and 5). Although each option is only learned over a single partition class, it can be initialized in any state.

{\bf Why partitions?}
If all options are trained on all data, then the options would not specialize defeating the purpose of learning multiple policies. Partitions are necessary to foster specialization. Natural partitionings arise in many different applications and are often easy to design. Consider navigation tasks (which we use in this paper for ease of visualization), which are ever-present in robotics \citep{Smart2002}, where partitions naturally lead an agent from one location to another in the state space. In addition, partitions are well suited to cyclical tasks; that is, tasks that have repeatable cycles (For example, a yearly cycle of 12 months). Here the state space can be easily partitioned based on time. Examples include inventory management systems \citep{Mann2014a} as well as maintenance scheduling of generation units and transmission lines in smart grids \citep{Abiri2013,Wu2010}.

{\bf Automatically Learning partitions}:
The availability of a pre-defined partitioning of the state space is a strong assumption in some domains. We have developed a relaxation to this assumption that can enable partitions to be learned \textit{automatically} using Regularized Option Interruption (ROI) \citep{Mann2014b, Sutton1999}.

{\bf Contributions}: Our main contributions are: \textbf{(1)} Introducing \Algorithm\ (\Alg), which learns options to repair and solve MPs. \textbf{(2)} Theorem \ref{thm:lsb} shows that \Alg\ converges to a near-optimal solution relating the quality of the learned policy to the quality of the options learned by the ``black box'' RL algorithm. \textbf{(3)} Theorem \ref{thm:ihomp_roi} proves that Regularized Option Interruption (ROI) can be safely incorporated into \Alg. \textbf{(4)} Experiments demonstrating that, given a misspecified problem, \Alg\ can learn options to repair and solve the problem. Experiments showing \Alg-ROI learning partitions and discovering reusable options. This divide-and-conquer approach may also enable us to scale and solve larger MDPs.

%%%%%%%%%%%%%%%%%%%%%%%%%%%%%%%%%%%%%%%%%%%%%%%%%%%%%%%%%%%%%%%%%%%%%%%%%%%%%%%
%
% BACKGROUND
%
%%%%%%%%%%%%%%%%%%%%%%%%%%%%%%%%%%%%%%%%%%%%%%%%%%%%%%%%%%%%%%%%%%%%%%%%%%%%%%%
\section{Background}

Let $M = \langle S, A, P, R, \gamma \rangle$ be an MDP, where $S$ is a (possibly infinite) set of states, $A$ is a finite set of actions, $P$ is a mapping from state-action pairs to probability distributions over next states, $R$ maps each state-action pair to a reward in $[0, 1]$, and $\gamma \in [0, 1)$ is the discount factor. A policy $\pi(a|s)$ gives the probability of executing action $a \in A$ from state $s \in S$.

Let $M$ be an MDP. The value function of a policy $\pi$ with respect to a state $s \in S$ is
$
V^{\pi}_{M}(s) = \mathbb{E} \left[ \sum_{t=1}^{\infty} \gamma^{t-1} R(s_t,a_t) | s_0 = s \right]
$ where the expectation is taken with respect to the trajectory produced by following policy $\pi$. The value function of a policy $\pi$ can also be written recursively as
\begin{equation} \label{eqn:value}
V^{\pi}_{M}(s) = \mathbb{E}_{a \sim \pi(\cdot|s)} \left[ R(s,a) \right] + \gamma \mathbb{E}_{s' \sim P(\cdot|s,\pi)} \left[ V^{\pi}(s') \right] \enspace ,
\end{equation}
which is known as the Bellman equation. The optimal Bellman equation can be written as
$
V^{*}_{M}(s) = \max_a \mathbb{E} \left[ R(s,a) \right] + \gamma \mathbb{E}_{s' \sim P(\cdot|s,\pi)} \left[ V^{*}(s') \right] \enspace .
$
Let $\varepsilon > 0$. We say that a policy $\pi$ is $\varepsilon$-optimal if $V^{\pi}_M(s) \geq V^{*}_M(s) - \varepsilon$ for all $s \in S$. The action-value function of a policy $\pi$ is defined by
$
Q^{\pi}_{M}(s,a) = \mathbb{E}_{a \sim \pi(\cdot|s)} \left[ R(s,a) \right] + \gamma \mathbb{E}_{s' \sim P(\cdot|s,\pi)} \left[ V^{\pi}(s') \right] \enspace ,
$
for a state $s \in S$ and an action $a \in A$, and the optimal action-value function is denoted by $Q^{*}_{M}(s, a)$. Throughout this paper, we will drop the dependence on $M$ when it is clear from the context.
\vspace{-0.2cm}
\section{Learning Options}
\vspace{-0.2cm}

An option is typically defined by a triple $o = \langle I, \pi, \beta \rangle$. However, we want to learn options that are both specialized to specific regions of the state-space but potentially reusable if they are useful in more general contexts. We focus on a special case of options, where an option $\sigma$ is defined by a tuple $\sigma = \langle \pi_\theta, \beta \rangle$, where $\pi_{\theta}$ is a parametric policy with parameter vector $\theta$ and $\beta : S \rightarrow \{ 0, 1\}$ indicates whether the option has finished ($\beta(s) = 1$) or not ($\beta(s) = 0$) given the current state $s \in S$.

Given a set of options $\Sigma$ with size $m \geq 1$, the inter-option policy is defined by $\mu : S \rightarrow [m]$ where $S$ is the state-space and $[m]$ is the index set over the options in $\Sigma$. An inter-option policy selects which options to execute from the current state by returning the index of one of the options. By defining inter-option policies to select an index (rather than the options), we can use the same policy even as the set of options is adapting.

Figure \ref{fig:compact}$b$ shows an arbitrary partitioning $\mathcal{P}$, consisting of $5$ sub-partitions $\{\mathcal{P}_i \vert i=1 \cdots 5\}$,  defined over the original MDP's state space. Each $\mathcal{P}_i$ is initialized with an arbitrary option and its corresponding \textbf{Local-MDP} $M_i'$. Local-MDP $M_i'$ (see supplementary material for a full definition) is an episodic MDP that terminates once the agent escapes from $\mathcal{P}_i$ and upon terminating receives a reward equal to the value of the state the agent would have transitioned to in the original MDP. Therefore, we construct a modified MDP called a Local-MDP and apply a planning or RL algorithm to solve it. The resulting solution (policy) is a specialized option.

Given a ``good'' set of options, planning can be significantly faster \citep{Sutton1999,Mann2014a}. However, in many domains we may not be given a good set of options. Therefore it is necessary to learn and improve this set of options. In the next section, we introduce an algorithm for dynamically learning and improving options using iterative hierarchical optimization.

%%%%%%%%%%%%%%%%%%%%%%%%%%%%%%%%%%%%%%%%%%%%%%%%%%%%%%%%%%%%%%%%%%%%%%%%%%%%%%%
%
% The Main Algorithm
%
%%%%%%%%%%%%%%%%%%%%%%%%%%%%%%%%%%%%%%%%%%%%%%%%%%%%%%%%%%%%%%%%%%%%%%%%%%%%%%%
\vspace{-0.2cm}
\section{\Algorithm\ (\Alg)}
\vspace{-0.2cm}

\Algorithm\ (\Alg, Algorithm \ref{alg:slb}) takes the original MDP $M$, a partition $\mathcal{P}$ over the state-space and a number of iterations $K \geq 1$ and returns a pair $\langle \mu, \Sigma \rangle$ containing an inter-option policy $\mu$ and a set of options $\Sigma$. The number of options $m = | \mathcal{P} |$ is equal to the number of classes (sub-partitions) in the partition $\mathcal{P}$
(line \ref{alg:slb:num_skills}). The inter-option policy $\mu$ returned by \Alg\ is defined
(line \ref{alg:slb:skill_policy})
by
$
%\begin{equation}
\mu(s) = \arg \max_{i \in [m]} \mathbb{I} \left\{ s \in \mathcal{P}_i \right\} \enspace ,
%\end{equation}
$
where $\mathbb{I} \{ \cdot \}$ is the indicator function returning $1$ if its argument is true and $0$ otherwise and $\mathcal{P}_i$ denotes the $i^{\rm th}$ class in the partition $\mathcal{P}$. Thus $\mu$ simply returns the index of the option associated with the partition class containing the current state. On line \ref{alg:slb:init_skills},
\Alg\ initializes $\Sigma$ with arbitrary options (\Alg\ can also be initialized with options that we believe might be useful to speed up learning).

\begin{algorithm}
\caption{\Algorithm\ (\Alg)}
\begin{algorithmic}[1]
\label{alg:slb}
\REQUIRE $M$\COMMENT{MDP}, $\mathcal{P}$\COMMENT{Partitioning of $S$}, $K$\COMMENT{Iterations}
\STATE $m \leftarrow | \mathcal{P} |$ \COMMENT{\# of partitions.} \label{alg:slb:num_skills}
\STATE $\mu(s) = \arg \max_{i \in [m]} \mathbb{I} \{ s \in \mathcal{P}_i \}$ \label{alg:slb:skill_policy}
\STATE Initialize $\Sigma$ with $m$ options. \COMMENT{1 option per partition.} \label{alg:slb:init_skills}
\FOR[Do $K$ iterations.]{$k = 1, 2, \dots, K$} \label{alg:slb:iters}
\FOR[One update per option.]{$i = 1, 2, \dots, m$} \label{alg:slb:updates}
\STATE \textbf{Policy Evaluation}:
\STATE Evaluate $\mu$ with $\Sigma$ to obtain $V^{\langle \mu, \Sigma \rangle}_{M}$ \label{alg:slb:evaluate}
\STATE \textbf{Option Update}:
\STATE Construct Local-MDP $M_{i}'$ from $M$ \& $V^{\langle \mu, \Sigma \rangle}_{M}$ \label{alg:slb:create_subgoal}
\STATE Solve $M_{i}'$ obtaining policy $\pi_{\theta}$ \label{alg:slb:solve_subgoal}
\STATE $\sigma_i' \leftarrow \langle \pi_{\theta}, \beta_i \rangle$ \label{alg:slb:update_skill}
\STATE Replace $\sigma_i$ in $\Sigma$ by $\sigma_i'$ \label{alg:slb:update_sigma}
\ENDFOR \label{alg:slb:updates_done}
\ENDFOR \label{alg:slb:iters_done}
\STATE \textbf{return} $\langle \mu, \Sigma \rangle$
\end{algorithmic}
\end{algorithm}

Next (lines \ref{alg:slb:iters}--\ref{alg:slb:iters_done}),
\Alg\ performs $K$ iterations. In each iteration, \Alg\ updates the options in $\Sigma$ (lines \ref{alg:slb:updates}--\ref{alg:slb:updates_done}).
Note that the value of a option depends on how it is combined with other options. If we allowed all options to change simultaneously, the options could not reliably propagate value off of each other. Therefore, \Alg\ updates each option individually. Multiple iterations are needed so that the option set can converge (Figure \ref{fig:compact}$b$).

The process of updating an option
(lines \ref{alg:slb:evaluate}--\ref{alg:slb:update_sigma})
starts by evaluating $\mu$ with the current option-set $\Sigma$
(line \ref{alg:slb:evaluate}).
Any number of policy evaluation algorithms could be used here, such as TD$(\lambda)$ with function approximation \citep{Sutton1998} or LSTD \citep{Boyan2002}, modified to be used with options. In our experiments, we used a straightforward variant of LSTD \citep{Sorg2010}. Then we use the original MDP $M$ to construct a Local-MDP $M'$ (line \ref{alg:slb:create_subgoal}).
Next, \Alg\ uses a planning or RL algorithm to approximately solve the Local-MDP $M'$ returning a parametrized policy $\pi_{\theta}$ (line \ref{alg:slb:solve_subgoal}).
Any planning or RL algorithm for regular MDPs could fill this role provided that it produces a parametrized policy. However, in our experiments, we used a simple actor-critic PG algorithm, unless otherwise stated. Then a new option $\sigma_i' = \langle \pi_\theta, \beta_i \rangle$ is created
(line \ref{alg:slb:update_skill})
where $\pi_\theta$ is the policy derived on line \ref{alg:slb:solve_subgoal}
and $\beta_i(s) = \left\{ \begin{array}{ll} 0 & \textrm{if } s \in \mathcal{P}_i \\ 1 & \textrm{otherwise} \end{array} \right.$. The definition of $\beta_i$ means that the option will terminate only if it leaves the $i^{\rm th}$ partition. Finally, we update the option set $\Sigma$ by replacing the $i^{\rm th}$ option with $\sigma_i'$
(line \ref{alg:slb:update_sigma}).
It is important to note that in \Alg, updating an option is equivalent to solving a Local-MDP.

%%%%%%%%%%%%%%%%%%%%%%%%%%%%%%%%%%%%%%%%%%%%%%%%%%%%%%%%%%%%%%%%%%%%%%%%%%%%%%%
%
% Analysis of IHOMP
%
%%%%%%%%%%%%%%%%%%%%%%%%%%%%%%%%%%%%%%%%%%%%%%%%%%%%%%%%%%%%%%%%%%%%%%%%%%%%%%%
\section{Analysis of \Alg}
\vspace{-0.1cm}
We provide the first convergence guarantee for combining hierarchically and iteratively learning options in a continuous state MDP using \Alg\ (Lemma 1 and Lemma 2, proven in the supplementary material). We use this guarantee as well as Lemma 2 to prove Theorem \ref{thm:lsb}. This theorem enables us to analyze the quality of the inter-option policy returned by \Alg. It turns out that the quality of the policy depends critically on the quality of the option learning algorithm. An important parameter for determining the quality of a policy returned by \Alg\ is the misspecification error defined below.

\begin{define}
\label{def:local}
Let $\mathcal{P}$ be a partition over the target MDP's state-space. The {\bf misspecification error} is
\begin{equation}
\eta_{\mathcal{P}} = \max_{i \in [m]} \eta_i \enspace ,
\end{equation}
where $\eta_i$ is the smallest $\eta_i \geq 0$, such that
$
V^{*}_{M_i'}(s) - V^{\pi_\theta}_{M_i'}(s) \leq \eta_i \enspace ,
$
for all $s \in \mathcal{P}_i$ and $\pi_\theta$ is the policy returned by the option learning algorithm executed on $M_i'$.
\end{define}

The misspecification error quantifies the quality of the Local-MDP solutions returned by our option learning algorithm. If we used an exact solver to learn options, then $\eta_{\mathcal{P}} = 0$. However, if we use an approximate solver, then $\eta_{\mathcal{P}}$ will be non-zero and the quality will depend on the partition $\mathcal{P}$. Generally, using finer grain partitions will decrease $\eta_{\mathcal{P}}$. However, Theorem \ref{thm:lsb} reveals that adding too many options can also negatively impact the returned policy's quality.

\begin{theorem} \label{thm:lsb}
Let $\varepsilon > 0$. If we run \Alg\ with partition $\mathcal{P}$ for $K \geq \log_{\gamma} \left( \varepsilon ( 1-\gamma ) \right)$ iterations, then the algorithm returns stitching policy $\varphi = \langle \mu, \Sigma \rangle$ such that
\begin{equation} \label{eqn:lsb_err}
\Vert V^*_{M} - V^{\varphi}_{M} \Vert_\infty \leq \frac{m\eta_{\mathcal{P}}}{(1-\gamma)^2} + \varepsilon \enspace ,
\end{equation}
where $m$ is the number of partition classes in $\mathcal{P}$.
\end{theorem}

The proof of Theorem \ref{thm:lsb} is divided into three parts (a complete proof is given in the supplementary material). The main challenge is that updating one option can impact the value of other options. Our analysis starts by bounding the impact of updating one option. Note that $\Sigma$ represents a option set and $\Sigma_i$ represents a option set where we have updated the $i^{th}$ option (corresponding to the $i^{th}$ partition class $\mathcal{P}_i$) in the set. In the first part, we show that the error between $ V_M^{*}$, the globally optimal value function, and $V_M^{\langle \mu, \Sigma_i \rangle}$, is a contraction when $s \in \mathcal{P}_i$ and is bound by $\Vert V_M^{*} - V_M^{\langle \mu, \Sigma \rangle} \Vert_\infty + \frac{\eta_{\mathcal{P}}}{1-\gamma}$ otherwise (Lemma 1).  In the second part, we apply an inductive argument to show that updating all $m$ options results in a $\gamma$ contraction over the entire state space (Lemma 2). In the third part, we apply this contraction recursively, which proves Theorem \ref{thm:lsb}.

This provides the first theoretical guarantees of convergence to a near optimal solution when combining hierarchically, and iteratively learning, a set of options $\Sigma$ in a continuous state MDP. Theorem \ref{thm:lsb} tells us that when the misspecification error is small, \Alg\ returns a near-optimal inter-option policy. The first term on the right hand side of (\ref{eqn:lsb_err}) is the approximation error. This is the loss we pay for the parametrized class of policies that we learn options over. Since $m$ represents the number of classes defined by the partition, we now have a formal way of analyzing the effect of the partitioning structure. In addition, complex options do not need to be designed by a domain expert; only the partitioning needs to be provided \textit{a-priori}. The second term is the convergence error. It goes to $0$ as the number of iterations $K$ increases.

The guarantee provided by Theorem \ref{thm:lsb} may appear similar to \citep[Theorem 1]{Hauskrecht1998}. However, \cite{Hauskrecht1998}  derive options only at the beginning of the learning process and do not update them. On the other hand, \Alg\ updates its option-set dynamically by propagating value throughout the state space during each iteration. Thus, \Alg\ does not require prior knowledge of the optimal value function.

Theorem \ref{thm:lsb} does not explicitly present the effect of policy evaluation error, which occurs with any approximate policy evaluation technique. However, if the policy evaluation error is bounded by $\nu > 0$, then we can simply replace $\eta_{\mathcal{P}}$ in (\ref{eqn:lsb_err}) with $(\eta_{\mathcal{P}} + \nu)$. Again, smaller policy evaluation error leads to smaller approximation error.

\section{Learning Partitions via Regularized Option Interruption}

So far \Alg\ has assumed a partition is given \textit{a-priori}. However, it may be non-trivial to design a partition and, in many cases, the partition may be sub-optimal. To relax this assumption, we incorporate Regularized Option Interruption (ROI) \citep{Mann2014b} into this work to enable \Alg\ to automatically \textit{learn} a near-optimal partition from an initially misspecified problem.

\Alg\ keeps track of the action value function $Q^{\langle \mu, \Sigma \rangle}(s,j)$ which represents the expected value of being in state $s\in S$ and executing option $j$, given the inter-option policy $\mu$ and option set $\Sigma$. ROI uses this estimate of the action-value function to enable the agent to choose when to switch options according to the following termination rule:
\begin{equation}
\beta_j(s,t)=\left\{ \begin{array}{ll}
1 & \textrm{if }Q^{\langle \mu, \Sigma \rangle}(s,j) < V^{\langle \mu, \Sigma \rangle}(s) -\rho\\
0 & \textrm{otherwise} \enspace .
\end{array}\right.
\label{eqn:zvalue_pi}
\end{equation}
\begin{wrapfigure}{r}{0.45\textwidth}
  \begin{center}
    \includegraphics[width=0.4\textwidth]{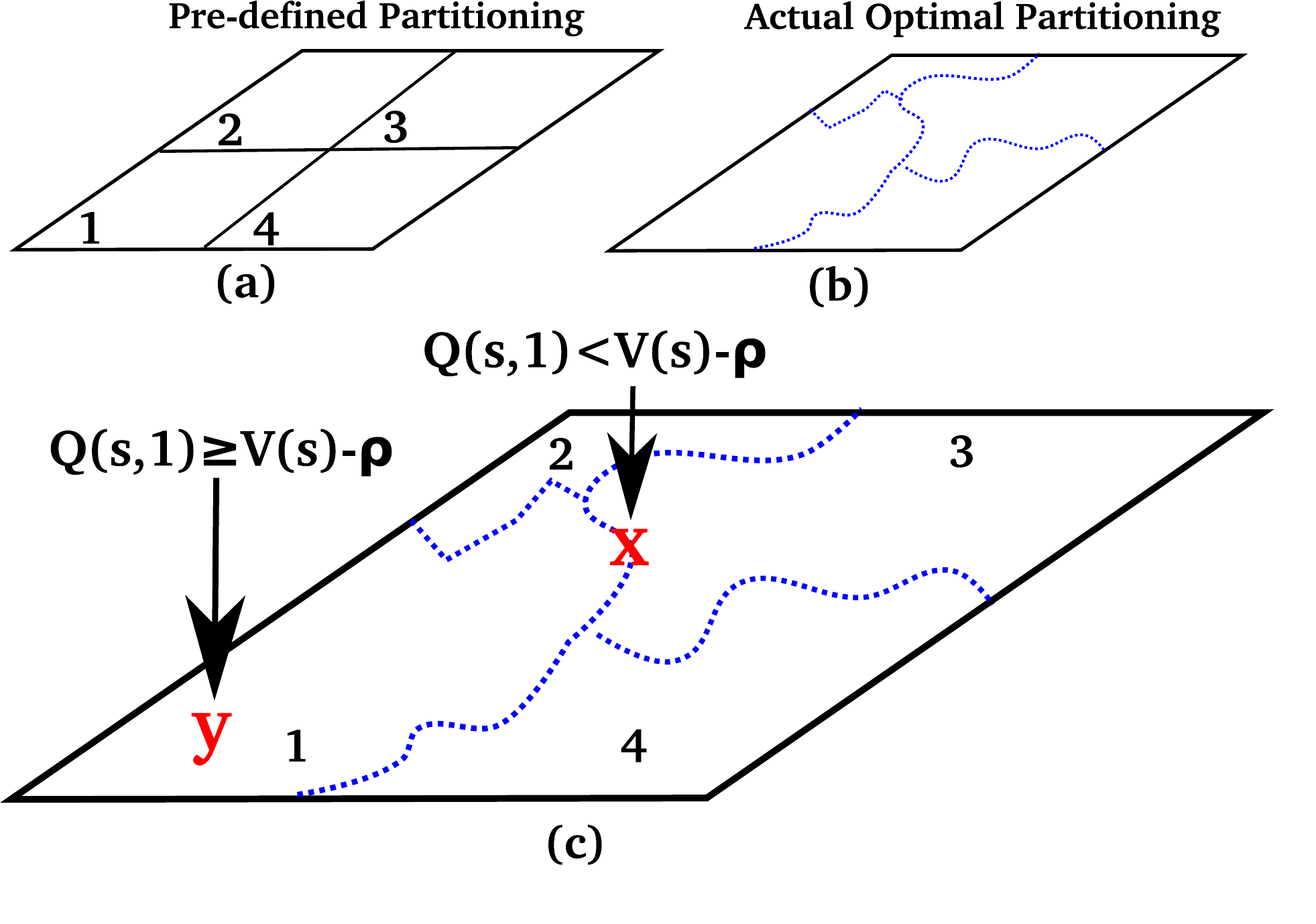}
  \end{center}
  \caption{Regularized Option Interruption (ROI): ($a$) The initial misspecified partition pre-defined by the user. ($b$) The actual optimal partitioning for the task. ($c$) The partition learned using ROI. If option $1$ is executing, the option will continue to execute at location $y$ and will terminate at location $x$. At this point option $3$ will begin to execute.}
  \label{fig:ptroi}
\end{wrapfigure}
Here $\beta_j(s,t)$ corresponds to the termination probability of the $j^{th}$ option partition and $V^{\langle \mu, \Sigma \rangle}(s) =\max_{i \in [m]} Q^{\langle \mu, \Sigma \rangle}(s,i)$. This rule is illustrated in Figure \ref{fig:ptroi}. A user has designed a partition resulting in a MP (Figure \ref{fig:ptroi}$a$) compared to the optimal partition for this domain (Figure \ref{fig:ptroi}$b$). \Alg\ applies ROI to `modify' the initial partition into the optimal one. By learning the optimal action-value function $Q^{\langle *, \mu, \Sigma \rangle}(s,j)$, \Alg\ builds a near-optimal partition (Figure \ref{fig:ptroi}$c$) that is implicitly stored within this action-value function. That is, if the agent is executing an option in partition class $1$, and the value of continuing with option $1$, $Q^{\langle \mu, \Sigma \rangle}(s,1)$, is less than $V^{\langle \mu, \Sigma \rangle}(s) -\rho$ for some regularization function $\rho$ (see the $x$ location in Figure \ref{fig:ptroi}$c$), then switch to the new option partition ($\beta_j(s,t)=1$). Otherwise, continue executing the current option (see the $y$ location in Figure \ref{fig:ptroi}$c$).

This leads to a new algorithm \Alg-ROI (\Alg\ with Regularized Option Interruption). The algorithm can be found in the supplementary material. The key difference between \Alg\ and \Alg-ROI is applying ROI during the policy evaluation step after each of the $m$ options have been updated. \Alg-ROI automatically learns an improved partition between iterations. We show that ROI can be safely incorporated into IHOMP in Theorem \ref{thm:ihomp_roi}. The theorem shows that incorporating ROI can only improve the policy produced by IHOMP. The full proof is given in the supplementary material.

\begin{theorem} \label{thm:ihomp_roi} {\bf (\Alg-ROI Approximate Convergence)}
  Eq. (\ref{eqn:lsb_err}) also holds for \Alg-ROI.
  %The result from Theorem \ref{thm:lsb} holds for \Alg-ROI.
  %Let $\varepsilon > 0$. If we run IHOMP-ROI with any admissible regularization function and initial partitioning $\mathcal{P}$ for $K \geq \log_\gamma(\varepsilon(1-\gamma))$ iterations, then the algorithm returns a stitching policy $\varphi$ such that
  %\begin{equation}
  %   \left\| V^*_M - V^\varphi_M \right\|_\infty \leq \left\| V^{*}_M - V^{\langle \mu,\Sigma\rangle}_M\right\|_\infty \leq \frac{m\eta_{\mathcal{P}}}{(1-\gamma)^2} + \varepsilon \enspace ,
  %\end{equation}
  %where $\langle \mu, \Sigma \rangle$ is the policy returned by \Algorithm\ without ROI, $m$ is the number of classes in $\mathcal{P}$ and $\eta_{\mathcal{P}}$ is the misspecification error.
\end{theorem}

%%%%%%%%%%%%%%%%%%%%%%%%%%%%%%%%%%%%%%%%%%%%%%%%%%%%%%%%%%%%%%%%%%%%%%%%%%%%%%%
%
% Experiments and Results
%
%%%%%%%%%%%%%%%%%%%%%%%%%%%%%%%%%%%%%%%%%%%%%%%%%%%%%%%%%%%%%%%%%%%%%%%%%%%%%%%
\section{Experiments and Results}
We performed experiments on three well-known RL benchmarks: Mountain Car (MC), Puddle World (PW) \citep{Sutton1996} and the Pinball domain \citep{Konidaris2009}. We also perform experiments in a sub-domain of \textit{Minecraft} \footnote{\url{https://minecraft.net/en/}}. The MC and Minecraft domains have similar results to PW and therefore have been moved to the supplementary material. We use two variations for the Pinball domain, namely \textit{maze-world} (moved to supplementary material), which we created, and \textit{pinball-world} which is one of the standard pinball benchmark domains.  Finally, we created a domain which we call the \textit{Two Rooms} domain to demonstrate how \Alg-ROI can improve partitions.

In each experiment, we defined a MP, where no flat policy is adequate, and in some of the tasks, cannot solve the task at all. These experiments simulate situations where the policy representation is constrained to avoid overfitting, manage system constraints, or coping with poorly designed features. In each case, \Alg\ learns a significantly better policy compared to the non-hierarchical approach. In the Two Rooms domain, \Alg-ROI improves the initial partition. Our experiments demonstrate potential to scale up to higher dimensional domains by hierarchically combining options over simple representations (A \textbf{Video} of \Alg\ solving the Pinball tasks and the Minecraft sub-domain can be found in the supplementary material).

\Alg\ is a meta-algorithm. We provide an algorithm for Policy Evaluation (PE) and Policy Learning (PL). For the MC and PW domains, we used SMDP-LSTD \citep{Sorg2010} for PE and a modified version of Regular-Gradient Actor-Critic (RG-AC) \citep{Bhatnagar2009} for PL (see supplementary material for details). In the Pinball domains, we used Nearest-Neighbor Function Approximation (NN-FA) for PE and UCB Random Policy Search (UCB-RPS) for PL. In the two rooms domain, we use a variation of LSTDQ with Option Interruption for PE and RG-AC for PL.

For the MC, PW, and Two Rooms domains, each intra-option policy is represented as a probability distribution over actions (independent of the state). We compare their performance to the original misspecified problem using a flat policy with the same representation. Grid-like partitions are generated for each task. Binary-grid features are used to estimate the value function.
In the Pinball domains, each option is represented by $5$ polynomial features corresponding to each state dimension and a bias term. The value function is represented by a KD-Tree containing $1000$ state-value pairs uniformly sampled in the domain. A value for a particular state is obtained by assigning the value of the nearest neighbor to that state that is contained within the KD-tree.
These are example representations. In principal, any value function and policy representation that is representative of the domain can be utilized.

\begin{figure}
\centering
  \includegraphics[width=0.70\textwidth]{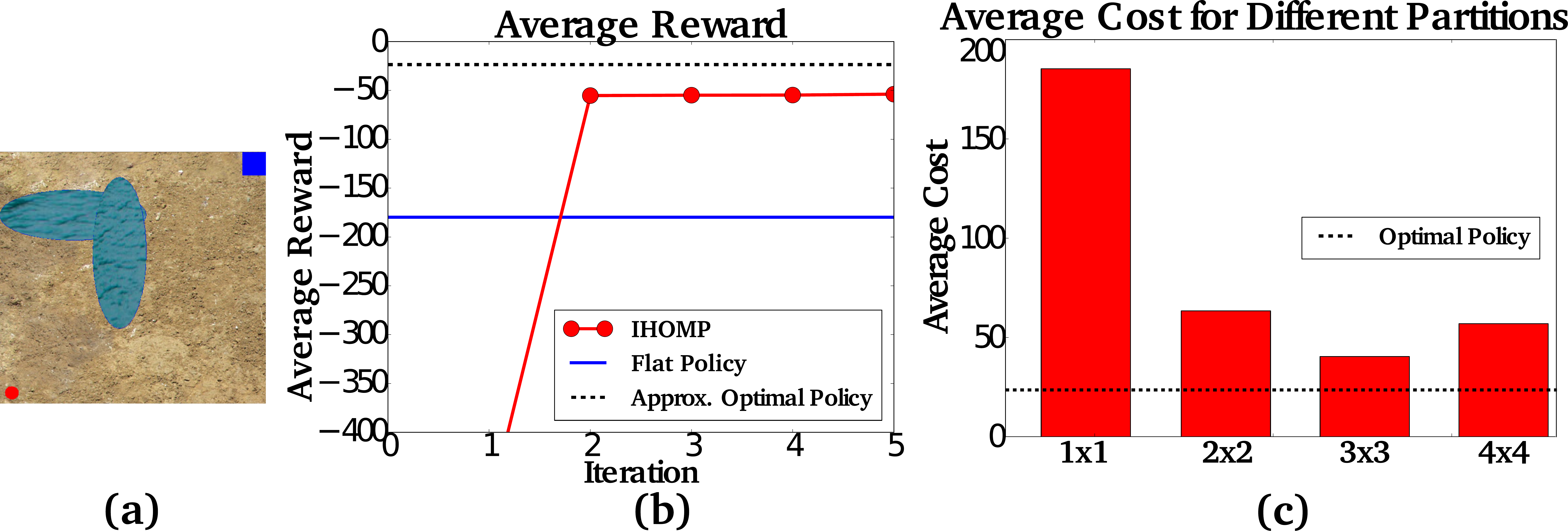}
\caption{($a$) Puddle World domain. ($b$) The average reward for \Alg compared to the solution obtained by a flat policy in the initially misspecified problem and an approximately optimal policy derived using Q-learning. ($c$) The average cost (negative reward) for each grid partition.}
\label{fig:pw}
\vspace{-0.7cm}
\end{figure}

%\subsection{Puddle World}
%\label{exp:pw}
{\bf Puddle World:}
Puddle World is a continuous 2-dimensional world containing two puddles as shown in Figure \ref{fig:pw}$a$. A successful agent (red ball) should navigate to the goal location (blue square), avoiding the puddles. The state space is the $\langle x,y \rangle$ location of the agent. Initially, the agent is provided with a misspecified problem. That is, a flat policy that can only move in a single direction (thus it cannot avoid the puddles). Figure \ref{fig:pw}$b$ compares this flat policy with \Alg\ (for a $2 \times 2$ grid partition (Four options)). The flat policy achieves low average reward. However, \Alg\ turns the flat policy into options and hierarchically composes these options together, resulting in a richer solution space and a higher average reward as seen in Figure \ref{fig:pw}$b$. This is comparable to the approximately optimal average reward attained by executing Approximate Value Iteration (AVI) for a huge number of iterations. In this experiment \Alg\ is not initiated in the partition class containing the goal state but still achieves near-optimal convergence after only $2$ iterations.

Figure \ref{fig:pw}$c$ compares the performance of different partitions where a $1 \times 1$ grid represents the flat policy of the initially misspecified problem. The option learning error $\eta_P$ is significantly smaller for all the partitions greater than $1\times 1$, resulting in lower cost. On the other hand, according to Theorem 1, adding more options $m$ increases the cost. A trade off therefore exists between $\eta_P$ and $m$. In practice, $\eta_P$ tends to dominate $m$. In addition to the trade off, the importance of the partition design is evident when analyzing the cost of the $3 \times 3$ and $4 \times 4$ grids. In this scenario, the $3 \times 3$ partition design is better suited to Puddle World than the $4 \times 4$ partition, resulting in lower cost.

\begin{figure}
\centering
\includegraphics[width=0.70\textwidth]{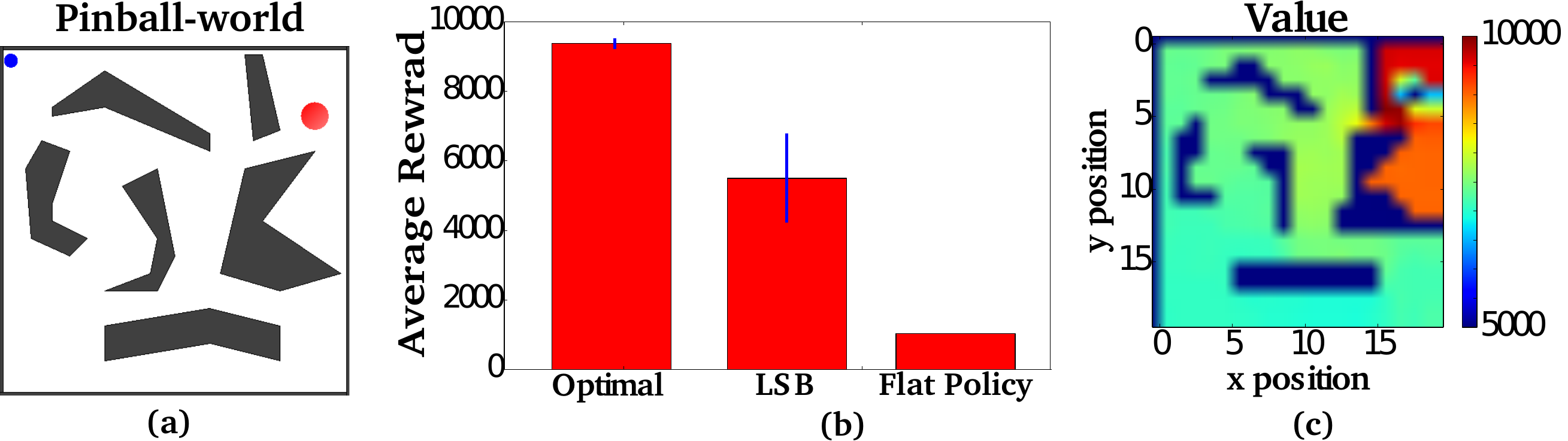}
\caption{($a$) Pinball domain. ($b$) Average reward for \Alg. \Alg\ converges after a single iteration as we start \Alg\ in the partition containing the goal. ($c$) The learned value function.}
\vspace{-0.3cm}
\label{fig:pinball_world}
\end{figure}

%\subsection{Pinball}
%\label{sec:pinball}
{\bf Pinball:}
We tested \Alg\ on the challenging pinball-world task (Figure \ref{fig:pinball_world}$a$) \cite{Konidaris2009}. The agent is initially provided with a $5$-feature flat policy $\langle 1, x, y, \dot{x}, \dot{y} \rangle$. This results in a misspecified problem as the agent is unable to solve the task using this limited representation as shown by the average reward in Figure \ref{fig:pinball_world}$b$. Using \Alg\ with a $4 \times 3 \times 1 \times 1$ grid, $12$ options were learned. \Alg\ clearly outperforms the flat policy as shown in Figure \ref{fig:pinball_world}$b$. It is less than optimal but still manages to sufficiently perform the task (see value function, Figure \ref{fig:pinball_world}$c$). The drop in performance is due to a complicated obstacle setup, non-linear dynamics and partition design. Nevertheless, this shows that \Alg\ can produce a reasonable solution with a limited representation.

%\subsection{Learning to improve the Partition}
%\label{sec:tworooms}
{\bf Improving Partitions:}
Providing a `good' option partitioning \textit{a-priori} is a strong assumption. It may be non-trivial to design the partitioning especially in continuous, high-dimensional domains. A sub-optimal partitioning may still mitigate misspecification, but it will not result in a near-optimal solution. To relax this assumption, we have incorporated Regularized Option Interruption into \Alg\ to produce the \Alg-ROI Algorithm. This algorithm learns the options and improves the \textit{partition}, effectively determining where the options should be executed in the state space.

\begin{figure}
\centering
\includegraphics[width=0.75\textwidth]{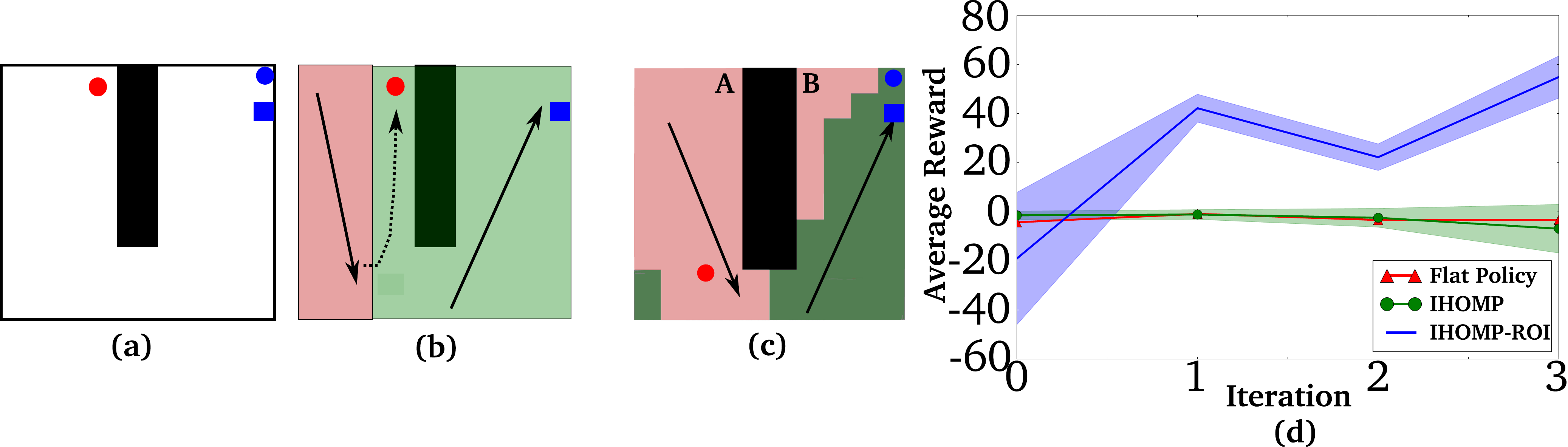}
\caption{($a$) Two Rooms domain ($b$) Two Rooms domain with a sub-optimal partition after performing \Alg\ ($c$) The learned partition after running \Alg-ROI ($d$) The average reward from executing \Alg-ROI compared to \Alg\ without ROI and the flat policy.}
\label{fig:tworooms}
\vspace{-0.5cm}
\end{figure}

We tested \Alg-ROI on the two rooms domain shown in Figure \ref{fig:tworooms}$a$. The agent (red ball) needs to navigate to the goal region (blue square). The policy parameterization is limited to a distribution over actions (moving in a single direction). This limited representation results in a MP as the agent is unable to traverse between the two rooms. If we use \Alg\ with a sub-optimal partitioning containing two options as shown by the red and green cells in Figure \ref{fig:tworooms}$b$, the problem is still misspecified. Here, the agent leaves the red cell and immediately gets trapped behind the wall whilst in the green cell. Using \Alg-ROI, as shown in Figure \ref{fig:tworooms}$c$, the agent learns both the options and a partition such that  the agent can navigate to the goal. The green region in the bottom left corner comes about as a function approximation error but does not prevent the agent from reaching the goal. If reader looks carefully in Figure \ref{fig:tworooms}$c$, they will notice something unexpected. The optimal partitioning learned for the Two Rooms domain includes executing the red option in region B. This is intuitive given the parameterizations learned for each of the options. The red option has a dominant right action whereas the green option has a dominant upward action. When the agent in region B, it makes more sense to execute the red option to reach the goal. Thus, \Alg-ROI provides an effective way to, not only learn an optimal partition, but to also discover where options should be reused.

%%%%%%%%%%%%%%%%%%%%%%%%%%%%%%%%%%%%%%%%%%%%%%%%%%%%%%%%%%%%%%%%%%%%%%%%%%%%%%%
% DISCUSSION
%%%%%%%%%%%%%%%%%%%%%%%%%%%%%%%%%%%%%%%%%%%%%%%%%%%%%%%%%%%%%%%%%%%%%%%%%%%%%%%
\vspace{-0.4cm}
\section{Discussion}
\vspace{-0.3cm}
We introduced \Alg\, a RL planning algorithm for iteratively learning options and an inter-option policy \citep{Sutton1999} to repair a MP. We provide theoretical results for \Alg\ that directly relate the quality of the final inter-option policy to the misspecification error. \Alg\ is the first algorithm that provides theoretical convergence guarantees while iteratively learning a set of options in a continuous state space. In addition, we have developed \Alg-ROI which makes use of regularized option interruption \citep{Sutton1999, Mann2014b} to learn an improved partition to solve an initially misspecified problem. \Alg-ROI is also able to discover regions in the state space where the options should be reused. In high-dimensional domains, partitions can be learned from expert demonstrations \citep{Abbeel2005} and intra-option policies can be represented as Deep Q-Networks \citep{Mnih2015}. Option reuse can be especially useful for \textit{transfer learning} \citep{Tessler2016} and multi-agent settings \citep{Garant2015}.

% Learning from Demonstration
%One limitation of \Alg\ is that it learns options for all partition classes. This may be a problem in high-dimensional state-spaces. However, the problem can be overcome, by focusing only on the most important regions of the state-space. One way to identify these regions is by observing an expert's demonstrations \cite{Abbeel2005,Argall2009}. In addition, we could apply self-organizing approaches to facilitate option reuse \cite{Moerman2009}. Option reuse can be especially useful for \textit{transfer learning}. Consider a multi-agent environment \cite{Garant2015} where many of the agents perform similar tasks requiring a similar option-set. Option reuse can also facilitate learning complex multi-agent policies (co-learning) with few samples.

\small
\bibliography{tmann}

\begin{thebibliography}{27}
\providecommand{\natexlab}[1]{#1}
\providecommand{\url}[1]{\texttt{#1}}
\expandafter\ifx\csname urlstyle\endcsname\relax
  \providecommand{\doi}[1]{doi: #1}\else
  \providecommand{\doi}{doi: \begingroup \urlstyle{rm}\Url}\fi

\bibitem[Abbeel and Ng(2005)]{Abbeel2005}
Pieter Abbeel and Andrew~Y Ng.
\newblock Exploration and apprenticeship learning in reinforcement learning.
\newblock In \emph{Proceedings of the 22nd International Conference on Machine
  Learning}, 2005.

\bibitem[Abiri-Jahromi et~al.(2013)Abiri-Jahromi, Parvania, Bouffard, and
  Fotuhi-Firuzabad]{Abiri2013}
Amir Abiri-Jahromi, Masood Parvania, Francois Bouffard, and Mahmud
  Fotuhi-Firuzabad.
\newblock A two-stage framework for power transformer asset maintenance
  management—part i: Models and formulations.
\newblock \emph{Power Systems, IEEE Transactions on}, 28\penalty0 (2):\penalty0
  1395--1403, 2013.

\bibitem[Bhatnagar et~al.(2009)Bhatnagar, Sutton, Ghavamzadeh, and
  Lee]{Bhatnagar2009}
Shalabh Bhatnagar, Richard~S Sutton, Mohammad Ghavamzadeh, and Mark Lee.
\newblock Natural actor--critic algorithms.
\newblock \emph{Automatica}, 45\penalty0 (11):\penalty0 2471--2482, 2009.

\bibitem[Boyan(2002)]{Boyan2002}
Justin~A Boyan.
\newblock Technical update: Least-squares temporal difference learning.
\newblock \emph{Machine Learning}, 2002.

\bibitem[Brunskill and Li(2014)]{Brunskill2014}
Emma Brunskill and Lihong Li.
\newblock {PAC}-inspired option discovery in lifelong reinforcement learning.
\newblock \emph{JMLR}, 2014.

\bibitem[Garant et~al.(2015)Garant, da~Silva, Lesser, and Zhang]{Garant2015}
Daniel Garant, Bruno~C. da~Silva, Victor Lesser, and Chongjie Zhang.
\newblock {Accelerating Multi-agent Reinforcement Learning with Dynamic
  Co-learning}.
\newblock Technical report, 2015.

\bibitem[Geramifard et~al.(2012)Geramifard, Tellex, Wingate, Roy, and
  How]{Geramifard2012}
A~Geramifard, S~Tellex, D~Wingate, N~Roy, and JP~How.
\newblock A bayesian approach to finding compact representations for
  reinforcement learning.
\newblock In \emph{European Workshops on Reinforcement Learning (EWRL)}, 2012.

\bibitem[Hauskrecht et~al.(1998)Hauskrecht, Meuleau, Kaelbling, Dean, and
  Boutilier]{Hauskrecht1998}
Milos Hauskrecht, Nicolas Meuleau, Leslie~Pack Kaelbling, Thomas Dean, and
  Craig Boutilier.
\newblock Hierarchical solution of markov decision processes using
  macro-actions.
\newblock In \emph{Proceedings of the 14th Conference on Uncertainty in AI},
  pages 220--229, 1998.

\bibitem[He et~al.(2011)He, Brunskill, and Roy]{He2011}
Ruijie He, Emma Brunskill, and Nicholas Roy.
\newblock Efficient planning under uncertainty with macro-actions.
\newblock \emph{Journal of Artificial Intelligence Research}, 40:\penalty0
  523--570, 2011.

\bibitem[Konidaris et~al.(2011)Konidaris, Osentoski, and Thomas]{Konidaris2011}
G.D. Konidaris, S.~Osentoski, and P.S. Thomas.
\newblock Value function approximation in reinforcement learning using the
  fourier basis.
\newblock In \emph{Proceedings of the Twenty-Fifth AAAI Conference on
  Artificial Intelligence}, 2011.

\bibitem[Konidaris and Barto(2009)]{Konidaris2009}
George Konidaris and Andrew~G Barto.
\newblock Skill discovery in continuous reinforcement learning domains using
  skill chaining.
\newblock In \emph{NIPS 22}, pages 1015--1023, 2009.

\bibitem[Levi and Weiss(2004)]{Levi2004}
Kobi Levi and Yair Weiss.
\newblock Learning object detection from a small number of examples: the
  importance of good features.
\newblock In \emph{Computer Vision and Pattern Recognition, 2004. CVPR 2004.
  Proceedings of the 2004 IEEE Computer Society Conference on}, volume~2, pages
  II--53. IEEE, 2004.

\bibitem[Mankowitz et~al.(2014)Mankowitz, Mann, and Mannor]{Mann2014b}
Daniel~J Mankowitz, Timothy~A Mann, and Shie Mannor.
\newblock Time regularized interrupting options.
\newblock \emph{ICML}, 2014.

\bibitem[Mann and Mannor(2014)]{Mann2014a}
Timothy~A Mann and Shie Mannor.
\newblock Scaling up approximate value iteration with options: Better policies
  with fewer iterations.
\newblock In \emph{Proceedings of the $\mathit{31}^{st}$ ICML}, 2014.

\bibitem[McGovern and Barto(2001)]{McGovern2001}
Amy McGovern and Andrew~G Barto.
\newblock {Automatic Discovery of Subgoals in Reinforcement Learning using
  Diverse Density}.
\newblock In \emph{Proceedings of the 18th ICML}, pages 361 -- 368, 2001.

\bibitem[Mnih(2015)]{Mnih2015}
Volodymyr et.~al. Mnih.
\newblock Human-level control through deep reinforcement learning.
\newblock \emph{Nature}, 2015.

\bibitem[Moerman(2009)]{Moerman2009}
Wilco Moerman.
\newblock \emph{Hierarchical reinforcement learning: Assignment of behaviours
  to subpolicies by self-organization}.
\newblock PhD thesis, Cognitive Artificial Intelligence, Utrecht University,
  2009.

\bibitem[Roy and How(2013)]{Roy2013}
N~Roy and JP~How.
\newblock A tutorial on linear function approximators for dynamic programming
  and reinforcement learning.
\newblock 2013.

\bibitem[Singh et~al.(1995)Singh, Jaakkola, and Jordan]{Singh1995}
Satinder~P Singh, Tommi Jaakkola, and Michael~I Jordan.
\newblock Reinforcement learning with soft state aggregation.
\newblock \emph{Advances in neural information processing systems}, pages
  361--368, 1995.

\bibitem[Smart and Kaelbling(2002)]{Smart2002}
William~D Smart and Leslie~Pack Kaelbling.
\newblock Effective reinforcement learning for mobile robots.
\newblock In \emph{Robotics and Automation, 2002. Proceedings. ICRA'02. IEEE
  International Conference on}, 2002.

\bibitem[Sorg and Singh(2010)]{Sorg2010}
Jonathan Sorg and Satinder Singh.
\newblock Linear options.
\newblock In \emph{Proceedings $9^{\rm th}$ AAMAS}, pages 31--38, 2010.

\bibitem[Sutton(1996)]{Sutton1996}
Richard Sutton.
\newblock Generalization in reinforcement learning: Successful examples using
  sparse coarse coding.
\newblock In \emph{Advances in neural information processing systems}, pages
  1038--1044, 1996.

\bibitem[Sutton and Barto(1998)]{Sutton1998}
Richard Sutton and Andrew Barto.
\newblock \emph{{Reinforcement Learning: An Introduction}}.
\newblock MIT Press, 1998.

\bibitem[Sutton et~al.(1999)Sutton, Precup, and Singh]{Sutton1999}
Richard~S Sutton, Doina Precup, and Satinder Singh.
\newblock {Between MDPs and semi-MDPs: A framework for temporal abstraction in
  reinforcement learning}.
\newblock \emph{AI}, 112\penalty0 (1):\penalty0 181--211, August 1999.

\bibitem[Tessler et~al.(2016)Tessler, Givony, Zahavy, Mankowitz, and
  Mannor]{Tessler2016}
Chen Tessler, Shahar Givony, Tom Zahavy, Daniel~J Mankowitz, and Shie Mannor.
\newblock A deep hierarchical approach to lifelong learning in minecraft.
\newblock \emph{arXiv preprint arXiv:1604.07255}, 2016.

\bibitem[Wu et~al.(2010)Wu, Shahidehpour, and Fu]{Wu2010}
Lei Wu, Mohammad Shahidehpour, and Yong Fu.
\newblock Security-constrained generation and transmission outage scheduling
  with uncertainties.
\newblock \emph{Power Systems, IEEE Transactions on}, 25\penalty0 (3):\penalty0
  1674--1685, 2010.

\bibitem[Zhou et~al.(2009)Zhou, Yuan, and Shi]{Zhou2009}
Huiyu Zhou, Yuan Yuan, and Chunmei Shi.
\newblock Object tracking using sift features and mean shift.
\newblock \emph{Computer vision and image understanding}, 113\penalty0
  (3):\penalty0 345--352, 2009.

\end{thebibliography}
\bibliographystyle{plainnat}

\end{document}